\documentclass[conference]{IEEEtran}
\IEEEoverridecommandlockouts
\usepackage{cite}
\usepackage{amsmath,amssymb,amsfonts}
\usepackage{graphicx}
\usepackage{textcomp}
\usepackage{xcolor}
\usepackage[T1]{fontenc}
\usepackage{cancel}
\usepackage{mathtools}
\usepackage{stmaryrd}
\usepackage{enumitem}
\usepackage{pifont}
\usepackage{xurl}
\usepackage{amsthm}
\usepackage{booktabs}
\usepackage{hyperref}
\usepackage{cleveref}

\usepackage[ruled,vlined,algo2e,linesnumbered]{algorithm2e}

\usepackage{amsmath,amsfonts,bm}

\def\eqref#1{equation~\ref{#1}}

\def\1{\bm{1}}

\def\va{{\bm{a}}}

\def\vh{{\bm{h}}}

\def\vx{{\bm{x}}}
\def\vy{{\bm{y}}}

\def\mA{{\bm{A}}}

\def\mH{{\bm{H}}}
\def\mI{{\bm{I}}}

\def\mW{{\bm{W}}}

\def\mTheta{{\bm{\Theta}}}

\DeclareMathAlphabet{\mathsfit}{\encodingdefault}{\sfdefault}{m}{sl}
\SetMathAlphabet{\mathsfit}{bold}{\encodingdefault}{\sfdefault}{bx}{n}

\def\gE{{\mathcal{E}}}

\def\gG{{\mathcal{G}}}

\def\gL{{\mathcal{L}}}

\def\gN{{\mathcal{N}}}

\def\gV{{\mathcal{V}}}

\def\gX{{\mathcal{X}}}
\def\gY{{\mathcal{Y}}}

\def\sG{{\mathbb{G}}}

\def\sR{{\mathbb{R}}}

\def\sY{{\mathbb{Y}}}

\newcommand{\name}{\textit{TADGNN}}

\makeatletter

\def\ps@IEEEtitlepagestyle{%
  \def\@oddfoot{\mycopyrightnotice}%
  \def\@evenfoot{}%
}
\def\mycopyrightnotice{%
  {\footnotesize 978-1-6654-8045-1/22/\$31.00 \copyright2022 IEEE\hfill}
  \gdef\mycopyrightnotice{}
}

\newcommand{\linebreakand}{%
  \end{@IEEEauthorhalign}
  \hfill\mbox{}\par
  \mbox{}\hfill\begin{@IEEEauthorhalign}
}
\makeatother

\newcommand{\etal}{et~al.\ }

\newcommand{\ie}{i.e.}

\newcommand{\xmark}{\ding{55}}%
\newcommand{\cmark}{\textcolor{orange}{\ding{51}}}%
\newcommand{\RNum}[1]{\uppercase\expandafter{\romannumeral #1\relax}}

\theoremstyle{definition}
\newtheorem{definition}{Definition}[section]

\begin{document}

\title{Dynamic Graph Node Classification via Time Augmentation}


\author{
\IEEEauthorblockN{Jiarui Sun\IEEEauthorrefmark{1}\thanks{\IEEEauthorrefmark{1}Work done while at Visa Research.}, Mengting Gu\IEEEauthorrefmark{2}, Chin-Chia Michael Yeh\IEEEauthorrefmark{2}, Yujie Fan\IEEEauthorrefmark{2}, Girish Chowdhary\IEEEauthorrefmark{3}, Wei Zhang\IEEEauthorrefmark{2}}

\IEEEauthorblockA{\IEEEauthorrefmark{1}Dept. of Electrical and Computer Engineering, \\ University of Illinois at Urbana, Champaign, Urbana, USA}
\IEEEauthorblockA{\IEEEauthorrefmark{3}Dept. of Computer Science and Dept. of Agricultural and Biological Engineering, \\ University of Illinois at Urbana, Champaign, Urbana, USA}
\IEEEauthorblockA{\IEEEauthorrefmark{2}Visa Research, Palo Alto, USA}
\{jsun57,girishc\}@illinois.edu, \{mengu,miyeh,yufan,wzhan\}@visa.com
}

\maketitle
\begin{abstract}
Node classification for graph-structured data aims to classify nodes whose labels are unknown.
While studies on static graphs are prevalent, few studies have focused on dynamic graph node classification.
Node classification on dynamic graphs is challenging for two reasons.
First, the model needs to capture both structural and temporal information, particularly on dynamic graphs with a long history and require large receptive fields.
Second, model scalability becomes a significant concern as the size of the dynamic graph increases.
To address these problems, we propose the Time Augmented Dynamic Graph Neural Network (\name) framework.
\textit{TADGNN} consists of two modules: 1) a time augmentation module that captures the temporal evolution of nodes across time structurally, creating a time-augmented spatio-temporal graph, and 2) an information propagation module that learns the dynamic representations for each node across time using the constructed time-augmented graph.
We perform node classification experiments on four dynamic graph benchmarks. Experimental results demonstrate that~\name~framework outperforms several static and dynamic state-of-the-art (SOTA) GNN models while demonstrating superior scalability.
We also conduct theoretical and empirical analyses to validate the efficiency of the proposed method.
Our code is available \href{https://sites.google.com/view/tadgnn}{\textit{here}}.
\end{abstract}


\section{Introduction}
Graph is a ubiquitous data structure that represents relationships between entities. 
Aiming to classify graph nodes whose labels are unknown, the task of node classification for graph-structured data has recently received increasing attention in various domains, such as biology~\cite{DBLP:conf/ijcai/XuCLL019} and social sciences~\cite{DBLP:journals/corr/abs-2006-10637}. 
This is attainable by the recent advance of Graph Neural Networks (GNNs), which generate node representations for classification through a message passing mechanism where each node iteratively aggregates neighborhood information. 

Existing GNNs mainly focus on static graphs~\cite{DBLP:conf/iclr/KipfW17,DBLP:conf/iclr/VelickovicCCRLB18}.
However, many real-world graphs are dynamic. 
Dynamic graphs can generally be categorized into discrete-time dynamic graphs and continuous-time dynamic graphs according to their representations.
Discrete-time dynamic graphs use an ordered sequence of graph snapshots, where each snapshot represents aggregated dynamic information within a fixed time interval.
Continuous-time dynamic graphs maintain detailed temporal information and are often more complex to model than the discrete case.
For both cases, graph structure, node attributes and node labels evolve over time in a complex manner.
Performing node classification on such dynamic graphs demands capturing this complicated evolving nature, which not only requires exploring the structural topology but also modeling the temporal aspect, which is absent in static GNN models.

In this work, we focus on node classification on discrete-time dynamic graphs. 
Because the existing dynamic graph learning methods are inefficient in both time and space as they either rely on recurrent structures\cite{DBLP:journals/pr/ManessiRM20} or the attention mechanism\cite{DBLP:conf/wsdm/SankarWGZY20}, these methods are not applicable for domains where dynamic graphs with a large number of time steps and nodes inhabit.
To this end, we propose a novel GNN framework named \textit{\textbf{\underline{T}}ime \textbf{\underline{A}}ugmented \textbf{\underline{D}}ynamic \textbf{\underline{G}}raph \textbf{\underline{N}}eural \textbf{\underline{N}}etwork} (\name).
\name~first employs a time augmentation module that constructs a time-augmented spatio-temporal graph based on the original graph snapshots. 
The time augmentation module \textit{efficiently} realizes the temporal evolution of nodes across time in a structural sense.
An information propagation module is then applied to \textit{effectively} learn the spatio-temporal information by aggregating node features from time-augmented neighborhoods.
Comparing with existing literature,~\name~achieves a good efficiency in two folds.
First, the time augmentation module models temporal graph dynamics without introducing significant computational demand since the time-augmented graph construction does not require learning.
In addition, by learning different importance scores for time-augmented neighborhoods of each node once, the dynamic attention mechanism introduced in the information propagation module enhances the model capacity while introducing little to no computational overhead.
These advantages make~\name~both powerful and efficient.
We summarize our key contributions as below:
\begin{itemize}[leftmargin=*]
\item A novel GNN framework named~\name~that works efficiently for node classification problems on dynamic graphs.
\item A theoretical and empirical analysis that validates the training efficiency of~\name. 
\item A comprehensive set of experiments that demonstrate the effectiveness of~\name~over SOTA methods.
\end{itemize}
\section{Related Work}

\subsection{Node Classification on Static Graphs}
Many works address static graph node classification problem by exploiting random walk statistics to optimize a stochastic measure of node similarities~\cite{DBLP:conf/kdd/GroverL16,DBLP:conf/kdd/PerozziAS14}.
Another line of research relies on graph spectral properties, designing convolutional filters based on the spectral graph theory~\cite{DBLP:journals/tsp/LevieMBB19,DBLP:journals/corr/abs-1901-01343}.
One significant work in this domain is Graph Convolutional Network (GCN)~\cite{DBLP:conf/iclr/KipfW17}, which restricts spectral filters to operate on each node's immediate neighbors, and can be interpreted as spatial aggregation.
It also inspires more spatial-based graph convolutional neural network models~\cite{DBLP:conf/nips/HamiltonYL17,DBLP:conf/iclr/VelickovicCCRLB18,DBLP:conf/iclr/XuHLJ19}. 

\subsection{Node Classification on Dynamic Graphs}

\subsubsection{Discrete-time Dynamic Graphs}

Many existing works utilize recurrent models for discrete-time dynamic graphs to capture the temporal dynamics into hidden states for classification. 
Some works use separate GNNs to model individual graph snapshot and use RNNs to learn temporal dynamics~\cite{DBLP:journals/pr/ManessiRM20}; 
some other works integrate GNNs and RNNs together into one layer, aiming to learn the spatial and temporal information concurrently \cite{DBLP:conf/aaai/ParejaDCMSKKSL20}. 
Sankar~\etal~\cite{DBLP:conf/wsdm/SankarWGZY20} use the self-attention mechanism along both the spatial and temporal dimensions of dynamic graphs, while Xu~\etal~\cite{DBLP:conf/ijcai/XuCLL019} combine the self-attention mechanism and RNNs together to learn the spatio-temporal contextual information jointly. 

\subsubsection{Continuous-time Dynamic Graphs}

Existing works on continuous-time dynamic graphs include RNN-based methods, temporal walk-based methods and temporal point process-based methods.
RNN-based methods perform node updates through recurrent models at fine-grained timestamps \cite{DBLP:conf/kdd/KumarZL19}, and the other two categories incorporate temporal dependencies through temporal random walks and parameterized temporal point processes \cite{DBLP:conf/www/NguyenLRAKK18,DBLP:conf/iclr/TrivediFBZ19}. 

\section{Preliminaries}


		
		
		


In this work, we focus on classifying nodes for discrete-time dynamic graphs $\sG= \{\gG^1, \gG^2, \ldots,  \gG^T\}$.
Each snapshot is an undirected graph $\gG^t = (\gV, \gE^t, \gX^t)$ at time step $t$. 
$\gV$ denotes the set of all the nodes appeared in $\sG$, and $|\gV| = N$ is the number of all nodes. 
$\gE^t \subseteq \gV^t \times \gV^t$ denotes the edge set of the snapshot at time $t$.
The edge set can also be represented as an adjacency matrix $A^t \in \sR^{N \times N}$ where $A^t_{uv} = 1$ if $(u, v) \in \gE^t$ otherwise $A^t_{uv}=0$.
$\gX^t \in \sR^{N \times d}$ denotes the node attribute matrix at time step $t$ where $d$ is the initial node feature dimension, and $\vx_v^t \in \sR^d$ is the feature vector of node $v$.
$\gY^t$ denotes the class label associated with each node at time step $t$, and $\vy_v^t$ is the class label of node $v$.
All nodes have their dynamic labels which evolve over time. 

\vspace{\baselineskip}
\noindent\textbf{Problem Statement:}
Given the dynamic graph history $\sG_{L} = \{\gG^1, \gG^2, \ldots,  \gG^t\}$ and the corresponding node labels $\sY_{L} = \{\gY^1,$ $\gY^2, \ldots, \gY^t\}$ up to time step $t$, we aim to classify all the nodes in future snapshots $\sG_{U}= \{\gG^{t+1}, \gG^{t+2}, \ldots,  \gG^T\}$ whose node labels $\sY_{U} = \{\gY^{t+1}, \gY^{t+2}, \ldots, \gY^T\}$ are unknown.

\begin{figure}[!tb]
\centering
\includegraphics[width=0.8\linewidth]{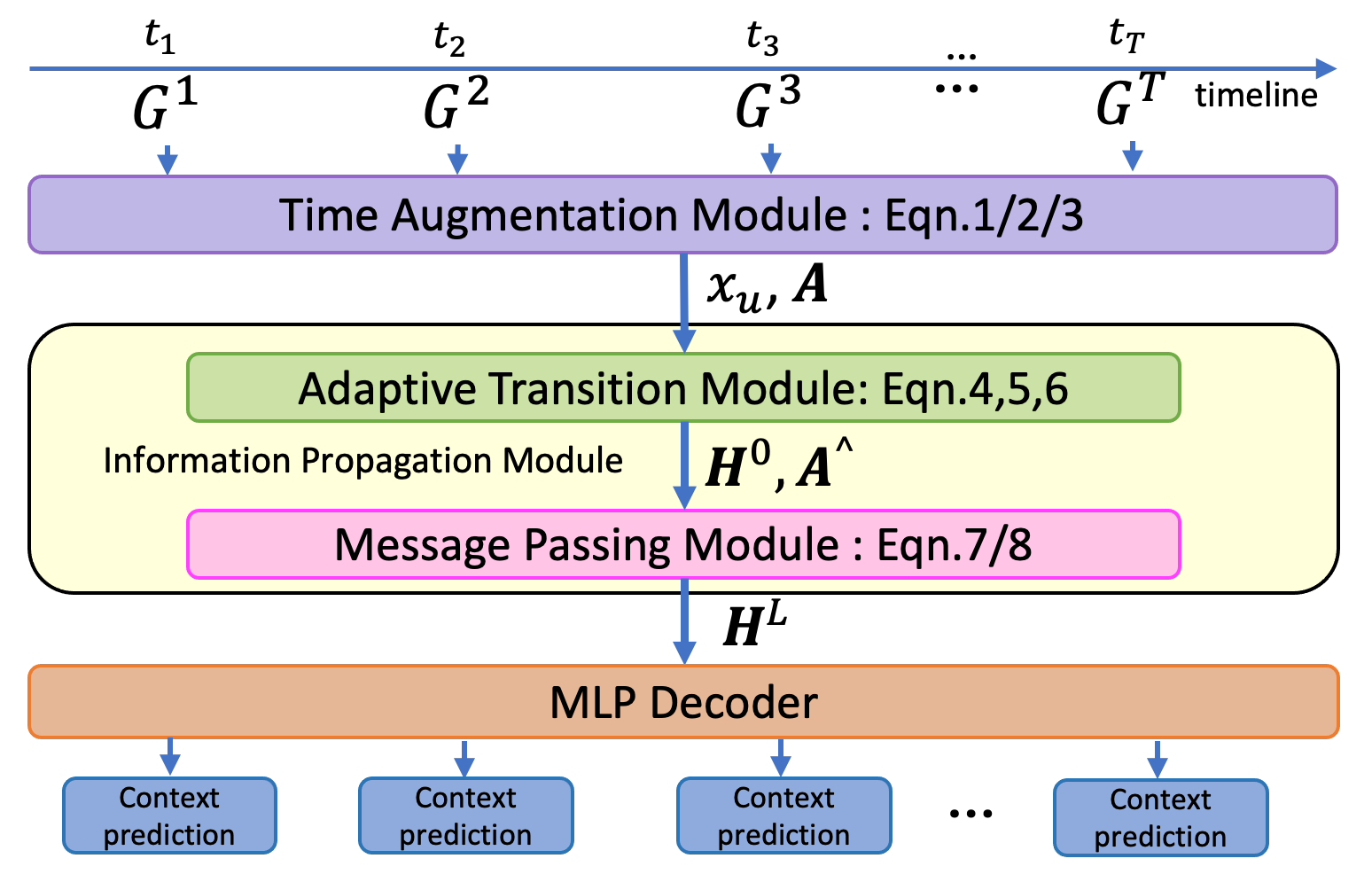}
\caption{\name~model architecture.} 
\label{fig:mdl}
\end{figure}
\section{\name~Architecture}
In this section, we present our framework, namely \textit{Time Augmented Dynamic Graph Neural Network} (\name), which is depicted in Fig. \ref{fig:mdl}.
We first introduce the proposed time augmentation module.
Then, we describe the information propagation module.
Finally, we discuss the time complexity of~\name~and compare it with several SOTA methods.
\subsection{Time Augmentation Module}
The time augmentation module is designed to model the temporal evolution of nodes across different snapshots in a structural sense.
To help explain our method, we define the \textit{temporal walk} under \textit{discrete-time dynamic graph} setting.
\begin{definition}[Temporal Walk]
Given a discrete-time dynamic graph sequence $\sG = \{\gG^1, \gG^2, \ldots,  \gG^T\}$, a \textit{temporal walk} from $v_1$ to $v_{k+1}$ is defined as $((v_1, v_2, t_1), (v_2, v_3, t_2)$ $, \ldots, (v_{k}, v_{k+1}, t_k))$ such that $(v_{i}, v_{i+1}) \in \gE^{t_i}$, $\gG^{t_i} = (\gV, \gE^{t_i})$ for $1 \leq i \leq k$, and $1 \leq t_1 \leq t_2 \leq \ldots \leq t_k \leq T$.
\end{definition}
A temporal walk is a time-respected walk that represents natural information flow through dynamic graph, but it is rarely explored in the literature.
To model such time-respected behaviors, we design the time augmentation module, which is inspired by the temporal PageRank algorithm~\cite{DBLP:conf/pkdd/RozenshteinG16}.
The time augmentation module aims to construct a time-augmented spatio-temporal graph, the walks on which can be considered as temporal walks simulated on the original dynamic graph. 
In this work, we consider three different realizations of such a time-augmented graph: the \textit{full time-augmentation realization}, the \textit{self-evolution time-augmentation realization}, and the \textit{disentangled time-augmentation realization}.
In the first two cases, the information propagation module is directly applied on the constructed time-augmented graph, propagating information structurally and temporally in a joint manner.
In the last case, the time-augmented graph is disentangled into a structural graph and a temporal graph, where different information propagation modules are applied on these two separate graphs as a two-stage process.
\subsubsection{Full Time-Augmentation Realization}
The first realization aims to produce a bijection between all possible walks on the time-augmented graph and all possible temporal walks on the original temporal graph.
Formally, we denote the adjacency matrix of the constructed time-augmented graph as $\mA \in \sR^{TN \times TN}$. Then, the adjacency matrix of the time-augmented graph can be represented as:
\begin{equation}
    \label{eq::ta1}
    \mA = \begin{bmatrix} 
    \tilde{A}^1 & A^2 & A^3 & \dots \\
    0 & \tilde{A}^2 & A^3 & \dots \\
    0 & 0 & \tilde{A}^3 & \dots \\
    \vdots & \vdots & \vdots & \ddots\\
    0 &  0 & 0      & \tilde{A}^T
    \end{bmatrix},
\end{equation}
where $\tilde{A}^t \in \sR^{N \times N}$ is the corresponding adjacency matrix at time step $t$ with self-loops added. 
\subsubsection{Self-Evolution Time-Augmentation Realization}
In the self-evolution time-augmentation realization, we restrict the cross-time walks such that only temporally adjacent nodes can be attended by their historical versions. Formally, let $v^t$ and $v^{t+1}$ denote node $v$ at time $t$ and $t+1$.
We only construct edge $(v^t, v^{t+1})$. 
By doing so, all temporal walks can be realized by performing random walk on the time-augmented graph as it can visit a node's neighbor at different time steps by visiting itself at different time steps first.
With previous notation, the adjacency matrix of the time-augmented graph can be represented as:
\begin{equation}
    \label{eq::ta2}
    \mA = \begin{bmatrix} 
    \tilde{A}^1 & I & 0 & \dots \\
    0 & \tilde{A}^2 & I & \dots \\
    0 & 0 & \tilde{A}^3 & \dots \\
    \vdots & \vdots & \vdots & \ddots\\
    0 &  0 & 0      & \tilde{A}^T
    \end{bmatrix},
\end{equation}
where $I \in \sR^{N \times N}$ is the identity matrix modeling node version updates across time. It is important to note that this realization of the time-augmented graph is relatively simplified and imposes trivial overhead in model space complexity.
\subsubsection{Disentangled Time-Augmentation Realization}
In the third case where we aim to completely disentangle structural and temporal modeling, we decouple the self-evolution time-augmentation case.
Instead of considering one time-augmented graph $\mA$, we decouple it into $\mA_{s} \in \sR^{TN \times TN}$ and $\mA_{t} \in \sR^{TN \times TN}$ as below:
\begin{equation}
    \label{eq::ta3}
    \mA_{s} = \begin{bmatrix} 
    \tilde{A}^1 & 0 & 0 & \dots \\
    0 & \tilde{A}^2 & 0 & \dots \\
    0 & 0 & \tilde{A}^3 & \dots \\
    \vdots & \vdots & \vdots & \ddots\\
    0 &  0 & 0      & \tilde{A}^T
    \end{bmatrix},
    \mA_{t} = \begin{bmatrix} 
    I & I & 0 & \dots \\
    0 & I & I & \dots \\
    0 & 0 & I & \dots \\
    \vdots & \vdots & \vdots & \ddots\\
    0 &  0 & 0 & I 
    \end{bmatrix}.
\end{equation}
%
\begin{table*}\scriptsize
\caption{Space and time complexity comparison between~\name~and other baselines.}
\centering
\begin{tabular}{cccc}
\toprule
Model Type & Space Complexity & Time Complexity & Sequential Ops. \\
\midrule
\name & $O(ET+LF^2+LTNF)$ & $O(LETF+LTNF^2)$ & $O(1)$ \\
DySAT & $O(LET+LF^2+LTNF+LNT^2)$ & $O(LETF+LTNF^2+LNT^2F)$ & $O(1)$\\
EvolveGCN & $O(ET+LTF^2+LTNF)$  & $O(LETF+LTNF^2)$ & $O(T)$\\
\bottomrule
\end{tabular}
\label{tab:complexity}
\end{table*}
%
\subsection{Information Propagation Module}
The information propagation module aims to capture both the structural and temporal properties of the dynamic graph via aggregating information from each node's time-augmented neighborhoods. 
We use $\gV_{ta}$, $\gE_{ta}$ to denote the node set and edge set of the time-augmented graph respectively, and $|\gV_{ta}| = N \times T$.
Thus, the input to the information propagation module is the initial node representations: $\{\vx_v^{t} \in \sR^{d}, \forall v^{t} \in \gV_{ta}\}$.
For convenience, we drop the superscript on node feature vectors and nodes as $\vx_v$ and $v$. 

\subsubsection{Adaptive Transition Module}

First, inspired by \cite{DBLP:journals/corr/abs-2105-14491}, we employ the dynamic attention mechanism to compute the adaptive information transition matrix, which allows different importance of nodes to be learned.
Formally, the adaptive graph transition matrix $\hat{\mA} \in \sR^{TN \times TN}$ is computed as:
\begin{equation}
\label{ATM}
e_{uv} = {\va}^T\sigma_{att}\left(\mA_{uv} \cdot \left(\left[\mTheta_L,\mTheta_R\right]\left[\vx_u||\vx_v\right]\right)\right),
\end{equation}
\begin{equation}
\label{ATMatt}
    \alpha_{uv} = \frac{\exp(e_{uv})}{\sum_{u \in \gN_v}\exp(e_{uv})},
\end{equation}
\begin{equation}
\label{ATM_output}
    \hat{\mA}^T_{uv} = \alpha_{uv},
\end{equation}
where $\mTheta_L \in \sR^{F \times d}$ and $\mTheta_R \in \sR^{F \times d}$ are the adaptive weight matrices applied on initial node features, $[\cdot,\cdot]$ is the horizontal concatenation of two matrix, $[\cdot||\cdot]$ is the vector concatenation operation, $\va \in \sR^{F}$ is a learnable weight vector for attention calculation, $\sigma_{att}(\cdot)$ is the LeakyReLU activation, $\gN_v=\{u \in \gV_{ta}: (u,v) \in \gE_{ta}\}$ denotes the immediate neighbor set of node $v$ on the time-augmented graph, and $\alpha_{uv}$ denotes the normalized attention coefficient of link $(u,v)$.
For the disentangled time-augmentation case, the adaptive structural and temporal transition matrices $\hat{\mA}_{s} \in \sR^{TN \times TN}$ and $\hat{\mA}_{t} \in \sR^{TN \times TN}$ are computed separately following the exact same procedures but with different parameters. 
We would like to emphasize that the dynamic attention mechanism we introduced does not impose quadratic computational complexity with respect to the number of snapshots as previously mentioned attention-based models \cite{DBLP:conf/wsdm/SankarWGZY20} since the attention scores are computed structurally for time-augmented neighborhoods.

\subsubsection{Message Passing Module}

In order to capture long range temporal and structural dependencies on the time-augmented graph, deep architectures are necessary. 
However, it is well-known that some GNN models such as GCN~\cite{DBLP:conf/iclr/KipfW17} and GAT \cite{DBLP:conf/iclr/VelickovicCCRLB18} demonstrate significant performance degradation when the number of layers increases \cite{DBLP:conf/aaai/LiHW18}. 
Thus, we adapt the message passing mechanism from GCNII~\cite{DBLP:conf/icml/ChenWHDL20} which relieves such over-smoothing issue through initial residual and identity mapping techniques. 
Formally, we denote the number of stacked information propagation layers as $L$ and the node representation output from the $l^{th}$ layer for the time-augmented graph as $\{\vh_v^{l} \in \sR^{F}, \forall v \in \gV\}$ where $F$ is the node representation dimension.
The input is either the initial encoded or intermediate node representations: $\{\vh_v^{l-1} \in \sR^{F}, \forall v \in \gV\}$.
When $l = 1$, \ie, at the first layer, we have $\vh_v^{0} = \mTheta_R \vx_v$.
Compactly, we have $\mH^l \in \sR^{TN \times F}$.
Then, we can define the message passing mechanism as: 
\begin{align}
\label{IPM}
\mH^{l+1}=\sigma_{ip}\Big(&\big((1-\alpha_{l}) \hat{\mA} \mH^{l}+\alpha_{l} \mH^{0}\big)\nonumber\\
&\big((1-\beta_{l}) \mI+\beta_{l} \mW^{l}\big)\Big),
\end{align}
where $\mW^{l} \in \sR^{F\times F}$ is the weight matrix, $\mI \in \sR^{F\times F}$ is the identity matrix, $\sigma_{ip}$ is the ReLU activation, $\alpha_{l}$ and $\beta_{l}$ are two hyperparameters adopting the same practice of GCNII~\cite{DBLP:conf/icml/ChenWHDL20}.
We also adapt GCNII variant, whose message passing mechanism is defined as:
\begin{align}
\label{IPM_variant}
\mH^{l+1}=\sigma_{ip}\Big(&(1-\alpha_{l}) \hat{\mA} \mH^{l}\big((1-\beta_{l})\mI+\beta_{l}\mW_{1}^{l}\big) +\nonumber\\
&\alpha_{l} \mH^{0}\big((1-\beta_{l}) \mI+\beta_{l} \mW_{2}^{l}\big)\Big),
\end{align}
where different weights $\mW_{1}^{l}, \mW_{2}^{l}$ are used for aggregated node representations $\hat{\mA} \mH^{l}$ and initial residual representations $\mH^{0}$.
Again, for the disentangled time-augmentation case, we compute the node representations $\mH$ by first utilizing structural transition matrix $\hat{\mA}_{s}$ to capture structural properties and then using temporal transition matrix $\hat{\mA}_{t}$ to learn temporal dynamics, with different weight matrices $\mW_{s}^{l}$ and $\mW_{t}^{l}$. 
With $\mA_{s}, \mA_{t}$ as the structural and temporal graph respectively, we first use one information propagation module for the structural graph $\mA_{s}$. 
After we obtain $\mH_s$ which summarizes structural information, we apply the second information propagation module guided by the temporal graph $\mA_{t}$, with $\mH_s$ as the initial node embedding input. 
Finally, we use $\mH^{L}$ that summarizes both structural and temporal dynamics for the dynamic node classification task.

\subsection{Learning Algorithm}
The outputs of the information propagation module $\mH^{L}$ are then feed into a multilayer perceptron (MLP) decoder which converts node representations to class logits, and are optimized to classify nodes in $\sG_{L}$ correctly as following:
\begin{equation}
\label{eq::loss}
\gL = \sum_{v\in\sG_{L}}J(y_v,\hat{y}_v), \hat{y}_v = \mathrm{MLP}(\vh_v^{L}),
\end{equation}
where $J(\cdot)$ measures weighted cross-entropy loss between groundtruth $y_v$ and predicted score $\hat{y}_v$. 
The loss weights are calculated based on class distribution from training partition.



\subsection{Complexity Analysis}
\label{sec::complexity}
In this section, we compare~\name's space and time complexity with DySAT and EvolveGCN, which can be considered as the representatives of attention-based and RNN-based dynamic graph learning models.
The space and time complexity for each method is shown in Table~\ref{tab:complexity}.
From the table, we can observe that comparing with both DySAT and EvolveGCN, \textit{TADGNN} has the lowest space complexity since DySAT is dominated by $O(LNT^2)$ term and EvolveGCN is dominated by $O(LTF^2)$ term.
In practice, memory space is a limiting factor for DySAT and EvolveGCN when $N$ and $L$ are necessarily large.
From the temporal perspective, the overall time complexity of~\name~is the lowest among the select baselines. DySAT's time complexity includes a $T^2$ term that makes it inefficient when modeling dynamic graphs with a large $T$.
As an RNN-incorporated GNN model, EvolveGCN has sequential operation dependence, which makes it infeasible to be processed in parallel and makes its practical training time significantly slower than purely convolution-based methods.

\section{Experiments}

In this section, we evaluate the effectiveness of \textit{TADGNN} for dynamic node classification task on four real-world datasets by comparing with four SOTA baselines.


\subsection{Datasets}
\begin{table}[t]\footnotesize
\caption{Dataset statistics}
\centering
\begin{tabular}{ccccccc}
\toprule
Datasets  & Nodes & Edges & Timestamps & Classes & Split\\
\midrule
\textsc{Wiki} & 9,227 & 2,833 & 11 & 2 & 4/3/4  \\
\textsc{Reddit} & 10,984 & 18,928 & 11 & 2 & 4/3/4  \\
\textsc{ML-Rating} & 9,746 & 90,928 & 11 & 5 & 4/3/4 \\
\textsc{ML-Genre} & 9,704 & 90,334 & 11 & 15 & 4/3/4  \\
\bottomrule
\end{tabular}
\label{tab:dataset}
\end{table}

We use four real-world dynamic graph datasets to conduct experiments, including two social networks, \textsc{Wiki}\cite{wiki_dataset} and \textsc{Reddit}\cite{reddit_dataset} and two rating networks, \textsc{ML-Rating}\cite{10.1145/2827872} and \textsc{ML-Genre}\cite{10.1145/2827872}.
All datasets are sliced into graph snapshot sequences.
Each snapshot contains information during fixed time intervals based on the timestamps provided in the raw data.
We also make sure that each snapshot contains sufficient interactions/links between nodes. 
Note that though these datasets are not attributed due to the difficulty of acquiring public attributed dynamic graphs,~\name~is designed for attributed dynamic graphs. 
As such, we use one-hot encoding of node IDs as node features in our experiments. Detailed dataset statistics are summarized in Table~\ref{tab:dataset}.

\subsection{Experimental Setup}


We select four SOTA node classification algorithms, including GAT \cite{DBLP:conf/iclr/VelickovicCCRLB18}, GCNII \cite{DBLP:conf/icml/ChenWHDL20}, EvolveGCN \cite{DBLP:conf/aaai/ParejaDCMSKKSL20} and DySAT \cite{DBLP:conf/wsdm/SankarWGZY20} to conduct model evaluation.
We use PyTorch to implement~\name~along with all other baselines.
We set the output feature dimension of the information propagation module to $128$. 
Adam~\cite{DBLP:journals/corr/KingmaB14}~is used as the optimizer with learning rate of $0.01$ along with weight decay of $0.0005$ as regularization to train all models for $200$ epochs in all experiments.
In particular, for~\name, we use validation set performance to tune the hyper-parameters and select the best time augmentation module type from three realizations. 
For baselines, we select hyper-parameters following the papers' tuning guidelines.
We report the averaged results along with corresponding standard deviations of five runs.
All models share the same MLP decoder architecture with dropout ratio $0.3$ to covert dynamic node representations to class logits.

\subsection{Node Classification Experiments} 
\label{sec:rq1}
In this section, we describe the conducted experiments and report the results together with the observed insights.

\subsubsection{Task Description}
Dynamic node classification task is used to evaluate~\name's effectiveness compared with other baselines.
All models are trained based on the training partition $\sG_{L} = \{\gG^1, \gG^2, \ldots, \gG^{t}\}$ with label information.
The task is to predict node labels in $\sG_{U} = \{\gG^{t+1}, \gG^{t+2}, \ldots, \gG^{T}\}$ by using the trained model with the entire dynamic graph $\sG = \{\gG^1, \gG^2, \ldots, \gG^{T}\}$ as input. Note that~\name~is inductive since it is agnostic to number of input graph snapshots.

\subsubsection{Experiment Setting}
Each dataset is sliced into a discrete graph snapshot sequence where each snapshot corresponds to a fixed time interval that contains a sufficient number of links. 
In each set of experiments, the first $t$ snapshots are used for model training. 
After training, we use the next $t'$ snapshots for validation, and the rest $T-t-t'$ snapshots for testing.
For all train/validation/test stages,~\name~processes input graph snapshots all at once.
We show the train/validation/test split statistics in Table~\ref{tab:dataset}.
Weighted cross-entropy loss is used as the objective function, that class weights obtained from training partition are used to relieve data imbalance issue.
The original DySAT model is temporally transductive as it cannot easily incorporate new graph snapshots for testing. Thus, we modify its position embeddings~\cite{DBLP:conf/icml/GehringAGYD17} to sine and cosine positional encodings as in Transformer~\cite{DBLP:conf/nips/VaswaniSPUJGKP17}.
For the static baselines, we employ one shared model across all snapshots.
All models are trained end-to-end.


\subsubsection{Evaluation Metric}
Since label information for all four datasets are highly imbalanced, we select \textit{Area Under the Receiver Operating Characteristic Curve} (AUC) metric to measure performance of different models. 
We use macro-AUC scores for evaluation.
Macro-AUC is computed by treating performances from all classes across all time steps (i.e., snapshots) equally, which is desirable for the case of dynamic node classification.

\begin{table}[t]\scriptsize
\caption{Dynamic node classification macro-AUC results with std.}
\centering
\begin{tabular}{lrrrrr}
\toprule
Model  &  \textsc{Wiki} &  \textsc{Reddit}  & \textsc{ML-Rating} & \textsc{ML-Genre}  \\
\midrule
GAT & $63.1 \pm 2.8$ & $62.9 \pm 1.0$ & $67.8 \pm 3.0$ & $\mathbf{81.2 \pm 2.6}$  \\
GCNII & $\underline{67.5 \pm 1.6}$ & $61.1 \pm 1.4$ & $\underline{71.5 \pm 3.7}$ & $62.1 \pm 1.2$ \\
\midrule
DySAT & $54.3 \pm 4.3$ & $\underline{66.5 \pm 0.3}$ & $53.9 \pm 5.8$ & $51.9 \pm 1.8$ \\
EvolveGCN & $58.9 \pm 2.1$ & $62.2 \pm 1.6$ & $61.3 \pm 2.5$ & $51.1 \pm 2.5$ \\
\name & $\mathbf{69.2 \pm 1.8}$ & $\mathbf{66.7 \pm 1.6}$ & $\mathbf{73.7 \pm 0.2}$ & $\underline{75.0 \pm 1.5}$ \\
\bottomrule
\end{tabular}
\label{tab:results_auc_new}
\end{table}

\begin{table}[t]\scriptsize
\caption{Ablation study of~\name~.}
\centering
\begin{tabular}{ccrr}
\toprule
\multicolumn{2}{c}{\textbf{Components}} & \multicolumn{2}{c}{\textbf{Perf.}}\\
\cmidrule(lr){1-2} \cmidrule(lr){3-4}
Time Augmentation & Adaptive Transition &  \multicolumn{1}{c}{{\textsc{Reddit}}} & \multicolumn{1}{c}{{\textsc{ML-Rating}}} \\
\midrule
\xmark & \xmark & $61.4 \pm 2.8$ & $71.2 \pm 0.6$\\
\cmark & \xmark & $63.8 \pm 0.8$ & $70.9 \pm 1.3$\\
\cmark & \cmark & $66.7 \pm 1.6$ & $73.7 \pm 0.2$\\
\bottomrule
\end{tabular}
\label{tab:ablation_macro}
\end{table}

\subsubsection{Results and Discussion}
We show the macro-AUC results in Table~\ref{tab:results_auc_new}.
Our observations include:
\begin{itemize}[leftmargin=*]
    \item \name~achieves superior performance on most of the datasets.
    This indicates that~\name~can better capture both structural and temporal graph dynamics comparing to other methods. On one dataset (\textsc{ML-Genre}) GAT performed well and has better result, but its performances have larger variance on different datasets.
    In addition, ~\name~is more efficient in space and time complexity, and can be applied in more scenarios especially where the dynamic graphs are large.

    \item Other dynamic baselines have inferior performance on certain datasets comparing to static methods. Besides, these dynamic baselines tend to keep larger variances for most datasets, which suggests that~\name~is more robust to random weight initialization. The results from our hyper-parameter search and analysis further indicate that the performance of these methods can be sensitive to hyper-parameter values; however, large time cost and space cost limit their potentials.
    This further suggests the importance of using temporal information efficiently to guide the dynamic graph node classification problem.
\end{itemize}

\subsection{Efficiency Comparison}
\label{sec:efficiency}
\begin{figure}[!tb]
\centering
\includegraphics[width=1.0\linewidth]{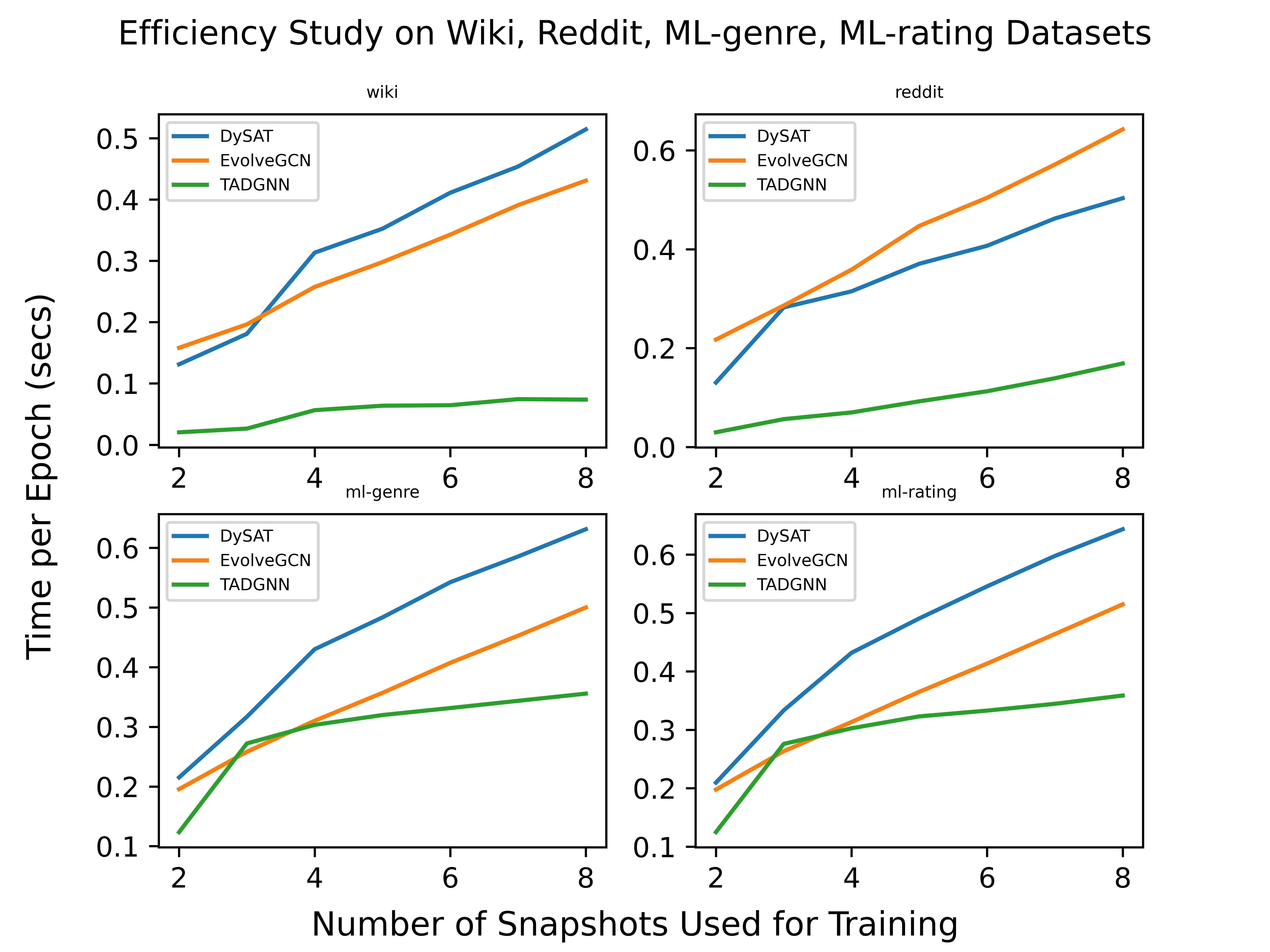}
\caption{Efficiency comparison among~\name, DySAT, and EvolveGCN.}
\label{fig:efficiency}
\end{figure}
In this section, we empirically demonstrate the efficiency advantage of~\name.
Specifically, we compare our model to EvolveGCN and DySAT on the average training time per epoch using different number of training snapshots.
These two methods are chosen since they represent widely used RNN-based and attention-based dynamic graph learning models.
For all three models, while keeping all common settings (\ie, learning rate) the same, we employ the best performing hyper-parameter setups. 
In particular, we compute the training time used per epoch averaged across $200$ epochs from $2$ to $8$ snapshots as training partition across all four datasets.

The efficiency comparison is shown in Fig. \ref{fig:efficiency}. 
The results are expected, as the training time of~\name~demonstrates a small increasing rate with respect to the number of training snapshots, while both DySAT and EvolveGCN depict much larger increasing rate as the number of training snapshots increases, due to the self-attention mechanism and sequential operation in RNN respectively. This empirical result confirms the efficiency advantage of~\name~over other deep dynamic graph learning methods.
More importantly, as the number of training snapshots and number of layers increase, both DySAT and EvolveGCN quickly fill up most of the GPU memories, thus hardly scalable to longer sequences or multi-layer setups, due to the extra memory requirements as discussed in Section \ref{sec::complexity}. In contrast,~\name~takes much less memory even when much more layers inhabit, which allows~\name~to capture long range temporal and structural properties when necessary. This empirical result validates our theoretical complexity analysis, demonstrating better efficiency of~\name, that it is powerful in modeling large dynamic graph datasets.

\subsection{Ablation Study} 
\label{rq3}

We conduct an ablation study to investigate how the time augmentation and information propagation modules affect~\name's modeling ability.
Specifically, we prepare three~\name~variants, \ie, 1) disable both the time augmentation and adaptive transition modules; 2) only disable the adaptive transition module; and 3) the full~\name~setup, and observe how disabling of different components affect the model performance.
We select two datasets (\textsc{Reddit} and \textsc{ML-Rating}) to cover different types of dynamic graphs.
We summarize our observations as below:
\begin{itemize}[leftmargin=*]
\item Both the time augmentation and adaptive transition modules are vital in temporal dynamics modeling, as they help~\name~achieve the best node classification results among different setups. In particular, we observe that our designed modules significantly boost~\name~performance on \textsc{Reddit}. The results suggest that temporal information is imperative to address dynamic node classification task.
\item Coupling the time augmentation and adaptive transition modules together helps~\name~reduce performance variance. We observe that enabling the time augmentation module only for \textsc{ML-Rating} increases performance variance significantly; on the contrary, though enabling time augmentation module only for \textsc{Reddit} reduces performance variance to the minimum, the performance can be further improved by introducing the adaptive learning module. This indicates the importance of distinguishing different levels of importance of node's time-augmented neighborhoods. 
\end{itemize}
\section{Conclusion}
Node classification on dynamic graphs has been gaining considerable attention. In this paper, we propose the Time Augmented Dynamic Graph Neural Network (\name) framework for discrete-time dynamic graph node classification. \name~consists of two modules: 1) a time augmentation module for converting the dynamic graph to a time-augmented spatio-temporal graph and 2) an information propagation module to process the time-augmented spatio-temporal graph. Evaluations are performed on four real-world dynamic graph benchmarks. Experimental results illustrate that \name~outperforms SOTA GNN models in terms of both accuracy and efficiency. 


\bibliographystyle{IEEEtran}
\bibliography{IEEEabrv, sample-base}

\end{document}